\documentclass[journal]{IEEEtran}

\usepackage{amsmath,amssymb}
\usepackage{graphicx}
\usepackage{booktabs}
\usepackage[colorlinks=true,linkcolor=blue,citecolor=blue,urlcolor=blue]{hyperref}

\newcommand{\maha}{maha}
\newcommand{\proto}{proto}
\newcommand{\cost}{cos\_t2}

\title{RSAJEV: Training-Free Abstaining Prototype Memory for Eliminating Silent Hallucination in JEV-Based Remote Sensing Classification}

\author{Man~Chen~\IEEEmembership{Member,~IEEE}, Antai~Chen, Shuai~Xu, Chengjie~Gu,\\
Huaping~Zhou, Kelei~Sun, Tao~Wu, and Yongjie~Huang%
\thanks{Manuscript received XXX; revised XXX. (Corresponding author: Antai Chen.)}
\thanks{Man Chen is with the School of Computer Science and Engineering, Anhui University of Science and Technology, Huainan 232001, Anhui, China, and also with the School of Electrical Automation and Information Engineering, Tianjin University, Tianjin 300072, China, and the State Key Laboratory of Digital Intelligent Technology for Unmanned Coal Mining, Huainan 232001, Anhui, China (e-mail: 2025110@aust.edu.cn).}%
\thanks{Antai Chen and Shuai Xu are with the School of Computer Science and Engineering, Anhui University of Science and Technology, Huainan 232001, Anhui, China (e-mail: chenantai\_hf@163.com; 2025201638@aust.edu.cn).}%
\thanks{Chengjie Gu, Huaping Zhou, Kelei Sun, and Tao Wu are with the School of Computer Science and Engineering, Anhui University of Science and Technology, Huainan 232001, Anhui, China, and also with the State Key Laboratory of Digital Intelligent Technology for Unmanned Coal Mining, Huainan 232001, Anhui, China (e-mail: cjgu@aust.edu.cn; hpzhou@aust.edu.cn; klsun@aust.edu.cn; taow@aust.edu.cn).}%
\thanks{Yongjie Huang is with the Guangxi Key Laboratory of Multi-Source Information Mining and Security, Guangxi Normal University, Guilin 541004, Guangxi, China.}}

\begin{document}
\maketitle

\begin{abstract}
Scene classification of remote sensing imagery with multimodal large language models (MLLMs) has recently been formulated as a single-forward-pass decision problem: candidate class names are rendered as lettered options in the prompt, a registered answer slot is conditioned on the full prompt, and classification is read out by scoring letter tokens with the language-modeling head. We show that although this letter binding is highly effective within its training vocabulary (97.71\% OA on UCM under the official protocol), it behaves erratically out of vocabulary: zero-adaptation accuracy across four cross-dataset directions fluctuates between 7.35\% and 96.40\%, determined purely by the training trajectory, and under out-of-vocabulary weights its confidence fails in the inverse direction on open-set evaluation, assigning higher confidence to unseen than to seen classes (AUROC 0.38)---the discriminative form of the generative failure mode we term \emph{silent hallucination}. We retain the slot representation and replace the readout with RSAJEV, a training-free abstaining prototype memory that performs retrieval-based prototype matching: $K$ reference images per class are cached as class prototypes via their slot representations under the same prompt, and classification becomes nearest-neighbor retrieval over the prototypes, refined by a shrinkage-Mahalanobis geometric correction. This readout is mechanistically isomorphic to retrieval-augmented generation (RAG) but operates on discriminative decision rather than text generation; to our knowledge, it is the first application of a retrieval-based readout to MLLM slot representations. The retrieval readout reaches 89.43\% on the UCM cross-dataset gate with 16 references per class and zero gradient updates, and 96.90\% on AID---the best zero-adaptation accuracy in this paper. Its ID-ness scores attain 0.91--0.99 AUROC on open-set evaluation over two datasets, exceeding symbolic confidence (0.38--0.96) in all four open-set settings, restoring a calibratable abstention channel to the generative interface and supporting the addition and removal of classes at zero extra inference cost through offline rescoring of the cached matrix.
\end{abstract}

\begin{IEEEkeywords}
Remote sensing scene classification, Multimodal large language models (MLLMs), silent hallucination, open-set recognition, prototype-based retrieval.
\end{IEEEkeywords}

\IEEEpeerreviewmaketitle

\section{Introduction}
\label{sec:intro}

\IEEEPARstart{M}{ultimodal} large language models (MLLMs)~\cite{gpt4,flamingo,llava} are increasingly used for remote sensing scene classification, where the task is cast as visual question answering over a fixed label vocabulary~\cite{geochat,lhrsbot}. A recent line of work~\cite{rsjev} compresses the entire decision into a single forward pass: instead of autoregressively generating a class name, the model appends a registered answer token to a prompt that lists all candidate classes, and reads out the classification from the hidden state at that slot position against the token rows of the language-modeling head. We call this family of methods \emph{one-pass deciders}. The design compresses a candidate vocabulary of up to 45 classes into a single forward pass, which is attractive for latency-sensitive remote sensing deployments; its accuracy under the official training protocol is 97.71\% OA on UCM. In this paper we additionally train UCM weights under the same official protocol as a second training source for the cross-dataset gate (Section~\ref{sec:closedset}).

The core of this design is letter binding, which deserves close scrutiny. Each class is bound to a letter token during slot-level fine-tuning; the class name itself enters the model only as prompt context. We measure two consequences. First, the binding fails to transfer across vocabularies, and the failure is erratic rather than gradual: AID-trained weights achieve only 21.52\% OA on UCM and 7.35\% on NWPU, whereas the same architecture self-trained on UCM reads an unseen 30-class AID vocabulary at 96.40\%; zero-adaptation behavior is determined by the training trajectory rather than vocabulary distance, and different checkpoints of the same mechanism differ by 89 points. Second, the binding leaves no exit for open-set inputs. Open-set inputs are the norm rather than the exception in deployment: the category list defined for one area rarely covers the next; ports, wind farms, and photovoltaic stations never appear in the source imagery yet arrive in the target region; land cover evolves over the model's service life; and annotation granularity can never exhaust every form the sensor will encounter. The candidate vocabulary defines the task, not the world. Under a 14-seen/7-unseen split of UCM, the maximum softmax, the top-1--top-2 margin, and the negative entropy of the letter distribution all separate seen from unseen classes in the wrong direction, with AUROC falling at 0.38--0.40: the readout is more confident about classes it has never seen than about those it has. This shares the fundamental flaw of its generative ancestors---the interface never abstains~\cite{hallsurvey}: whether or not the true class is among the candidates, the trained readout outputs a sharp distribution over the listed options, and the model masks not-knowing with confident answers. This is the discriminative form of the silent hallucination that plagues generative MLLMs, and it is why a classification interface must be paired with a readout whose confidence actually means something. With clean training weights that have never seen the out-of-distribution (OOD) vocabulary, the inversion disappears (0.76), but the symbolic score still ranks behind prototype similarity. On the checkpoint with the weakest letter head, the same slot representation can still be decoded by a training-free Mahalanobis readout to near the level of the cross-dataset gate (Table~\ref{tab:gate}, in-domain rows). The variance sits in the symbolic head and the training trajectory, not in the slot representation that this paper's readout consumes.

Both failures reside in the readout, not the representation. We therefore keep the slot representation intact and replace the readout with retrieval-based, training-free prototype matching, refined by a shrinkage-Mahalanobis readout; we call the resulting method RSAJEV. This readout is isomorphic to retrieval-augmented generation (RAG): the slot representations of $K$ reference images per class are cached as class prototypes, classification is nearest-neighbor retrieval in the same space, and no gradient update is involved; to our knowledge, this mechanism is applied to MLLM slot representations for the first time. Symbolic scoring and cosine matching participate only as control readouts. With 16 references per class and zero gradient updates, the retrieval readout gains more than 10 points over the cosine prototype before whitening on the UCM cross-dataset gate and achieves the best zero-adaptation accuracy in this paper on AID; the symbolic readout, by contrast, swings wildly across the four directions depending on the shipped weights, with its training trajectory the only determining factor (Section~\ref{sec:gate}, Table~\ref{tab:gate}). We choose shrinkage Mahalanobis rather than stopping at cosine because of the anisotropy of slot space: prototype matching implies an isotropy assumption, whereas slot-level fine-tuning pulls the representation geometry toward the letter-readout direction, leaving a systematic deformation that can be whitened by a covariance estimated from the references; the shrinkage coefficient suppresses the ill-conditioned condition number of small-sample covariance estimation. In open-set evaluation the ranking is consistent: the retrieval readout attains 0.91--0.99 AUROC, symbolic confidence falls between 0.38 and 0.96, and ranks behind the retrieval readout in every setting. A column-pruning experiment further shows that the slot representation is prompt-robust: shrinking the candidate set from 21 to 14 classes moves the seen-class accuracy of the text and prototype readouts by at most 3.7 points, so the difference stems from the readout rather than from representation drift.

Our contributions are threefold. First, we dissect the one-pass decider into a shared representation and an interchangeable readout, and quantify the prompt robustness of the representation with a pruning experiment that requires no extra forward passes. Second, we propose RSAJEV, the retrieval-based, training-free prototype readout: slot representations of $K$ reference images per class are cached as prototypes, classification is nearest-neighbor retrieval, and shrinkage-Mahalanobis geometry refines the metric; to our knowledge, this is the first retrieval-based readout on MLLM slot representations. The readout characterizes a clean data--compute trade-off: at the total cost of $K$ references per class and one shared forward pass, it achieves the best zero-adaptation accuracy in this paper on two target vocabularies, whereas the fate of symbolic transfer is decided by the training trajectory (Section~\ref{sec:gate}). Third, we show that symbolic confidence is unusable for open-set rejection: negatively correlated with novelty under out-of-vocabulary weights and still weaker than prototype similarity under clean weights---that is, the one-pass decider inherits the silent hallucination of its generative ancestor in discriminative form; replacing the readout restores the abstention channel, and the abstention operating point can be flexibly set via an ID-ness threshold to match different false-positive budgets. All readouts share one cached forward pass per image; open-set rejection, vocabulary changes, and class additions all reduce to offline rescoring.

\section{Related Work}
\label{sec:related}

\begin{figure*}[!t]
  \centering
  \includegraphics[width=\textwidth]{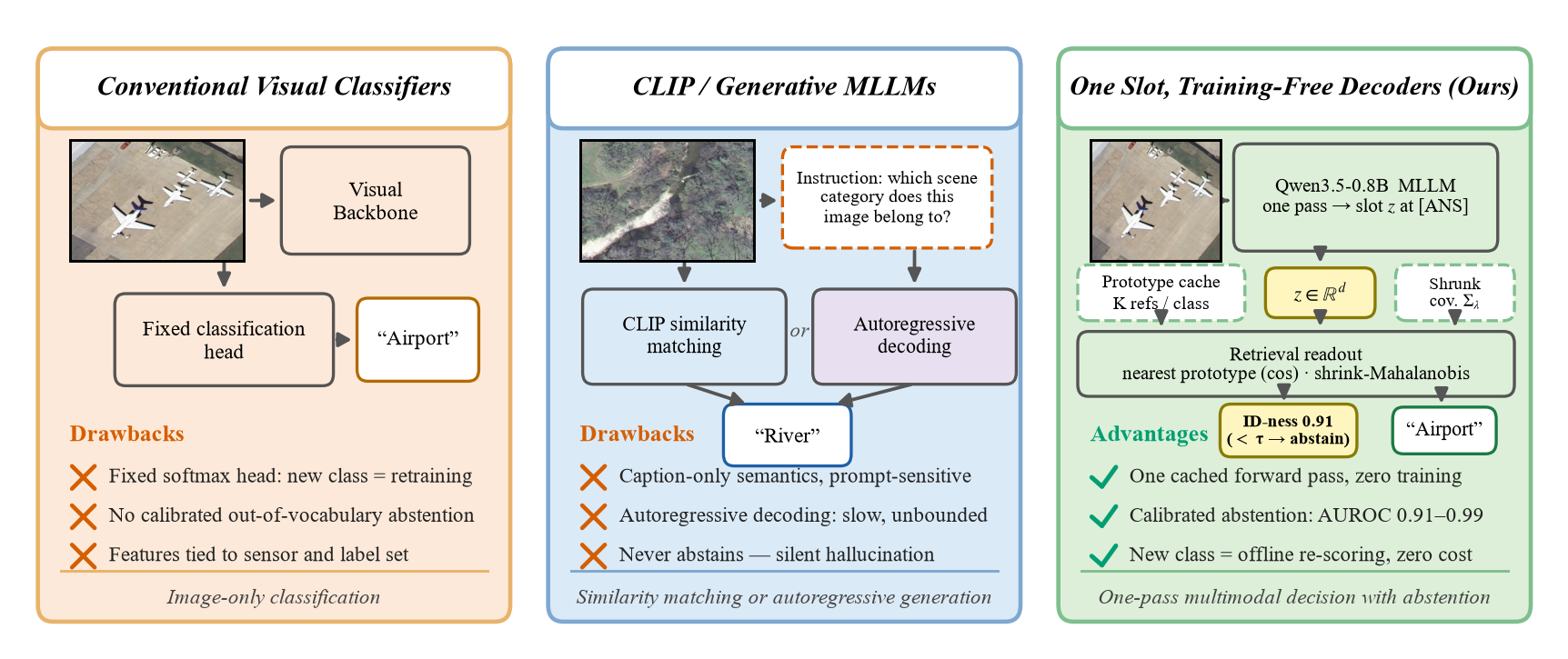}
  \caption{Comparison of three families of methods: structural differences, strengths, and weaknesses of conventional visual classifiers, CLIP/generative MLLMs, and this paper's approach. Our column is the main line of the method; the symbolic and text readouts serve only as controls.}
  \label{fig:families}
\end{figure*}

\paragraph{Task-specific vision models.}
Supervised convolutional networks since ImageNet~\cite{imagenet}---from AlexNet~\cite{alexnet} and VGG~\cite{vgg} to ResNet~\cite{resnet}---together with subsequent Transformer backbones~\cite{vit,swin} and self-supervised pretraining~\cite{dino,mae} defined the first generation of remote sensing scene classifiers, which remain the reference frame for in-domain accuracy. Their displacement is motivated by three structural defects: the classifier is a softmax over the training vocabulary, so adding or removing a class means retraining the head; softmax scores have no calibration mechanism for out-of-vocabulary inputs, so a closed-set model cannot abstain; and the learned features are bound to the training sensor and label set, so neither classifier nor representation transfers without supervision. These are exactly the faces measured one by one in Section~\ref{sec:experiments}.

\paragraph{Vision--language pretraining.}
CLIP~\cite{clip} aligns an image encoder with a text encoder contrastively, performs classification by scoring class-name descriptions, removes the retraining requirement, and achieves zero-shot transfer to unseen vocabularies; remote sensing derivatives~\cite{remoteclip} adapt this pairing to aerial imagery and narrow the domain gap. On standard scene benchmarks, however, zero-shot accuracy still lags far behind supervised and MLLM pipelines~\cite{rsjev}, results are sensitive to prompt wording, and class semantics reside entirely in generic description text---the image representation is never conditioned on the candidate list of the question at hand. Our text control readout is the nearest relative of this paradigm in our setting, likewise scoring class names against text; its measurements draw an upper bound for the paradigm from the representation side: slot training pulls the last-layer geometry away from description alignment, and the text readout drops to 38.10 in-domain on UCM (Table~\ref{tab:gate}), far below the 78.10 symbolic readout on the same slot representation.

\paragraph{Generative MLLMs and silent hallucination.}
Remote sensing MLLMs~\cite{geochat,lhrsbot,earthgpt,rsgpt,h2rsvlm} formulate scene classification as visual question answering and autoregressively decode a class name; their capability ceiling is set by the general-purpose backbones~\cite{gpt4,qwenvl,qwen25vl,qwen3vl}. This interface inherits two problems from generation: latency grows with answer length, and the output space is unbounded---the model may emit labels outside the candidate set, or refuse to commit. More fundamentally, the model always produces a fluent answer with no calibrated channel toward ``unknown''~\cite{hallsurvey}: when the model does not know, it does not abstain but hallucinates, a failure mode systematically documented in large language and vision--language models. H2RSVLM~\cite{h2rsvlm} likewise treats ``honesty'' as a standalone objective in the remote sensing setting, confirming that the problem is not domain-specific. Silent hallucination denotes fluent, confident answering with no mechanism signaling that the model does not know---the most destructive error mode in classification, because a confident out-of-vocabulary answer is indistinguishable from a correct one downstream. RSJEV~\cite{rsjev} compresses classification into a one-pass lettered-option decision to cut latency; our measurements show that the confidence problem survives compression: the letter head always outputs a sharp distribution over the listed options, ``none of the above'' is unreachable by construction, and its confidence can even be negatively correlated with novelty (AUROC 0.38, Table~\ref{tab:ucmoov})---the discriminative form of silent hallucination. This paper keeps the representation that makes the one-pass decider fast and replaces the readout that makes it unaccountable.

\paragraph{Open-set recognition and training-free adaptation.}
Open-set scene classification requires rejecting inputs from classes outside the training vocabulary~\cite{bendale2016}, and progress has been systematically surveyed in~\cite{opensetsurvey}. Standard protocols report AUROC and OSCR of ID-ness scores (the latter couples rejection with closed-set correctness). Threshold-based post-hoc detectors form a long lineage: the maximum-softmax baseline~\cite{msp}, ODIN~\cite{odin}, the energy score~\cite{energy}, likelihood ratios~\cite{likelihoodratio}, and the class-conditional Mahalanobis score~\cite{mahalanobis}; the latter is geometrically of the same origin as our shrinkage-Mahalanobis readout but operates in a classifier's feature space---we transplant the shrinkage covariance to MLLM slot space and re-estimate its small-sample conditioning. A recurring pitfall of remote sensing benchmarks is pseudo-open-set construction, where classes from related datasets serve as unknowns although the underlying model has already seen them; splits within a single dataset avoid this confound. Our open-set protocol uses within-dataset construction and adds an easily overlooked failure mode: the option-token head of discriminative training can be negatively correlated with novelty, which condemns threshold-based rejection before any threshold is chosen. On the adaptation side, cache-based training-free adapters---matching networks~\cite{matchingnets}, relation networks~\cite{relationnet}, MAML~\cite{maml}, few-shot prototypical networks~\cite{protonet}, and Tip-Adapter~\cite{tipadapter}---replace gradient updates with a lightweight memory of reference features. Our retrieval-based prototype readout applies this cache-adaptation principle to MLLM slot representations rather than visual-encoder features; to our knowledge, this is the first time the principle enters MLLM slot space. It brings two setting-specific changes: reference representations must be computed under the same option-rendered prompt as inference; and because the representation is prompt-robust (which we verify experimentally), the resulting cache can be reused across candidate-set changes.

\paragraph{Positioning.}
Against these three families, the ``slot plus readout'' perspective gives one consistent path. Relative to task-specific vision models, we remove the retrain-on-vocabulary-change requirement while keeping MLLM-grade accuracy (Table~\ref{tab:gate}: 89.43\% on UCM with 16 references per class and zero gradients). Relative to CLIP-style description scoring, we replace generic description text with visual references in the same space, reaching 89.43\% cross-dataset and 96.90\% on AID---the best zero-adaptation accuracy in this paper. Relative to generative and one-pass MLLMs, we supply the abstention channel that both lack, making ``does the model know'' a measurable score (0.91--0.99 AUROC on two datasets) rather than a property of answer fluency. Figure~\ref{fig:families} juxtaposes the structural limitations of the three families and the abstention channel that our readout restores.

\section{Method}
\label{sec:method}

\begin{figure*}[!t]
  \centering
  \includegraphics[width=\textwidth]{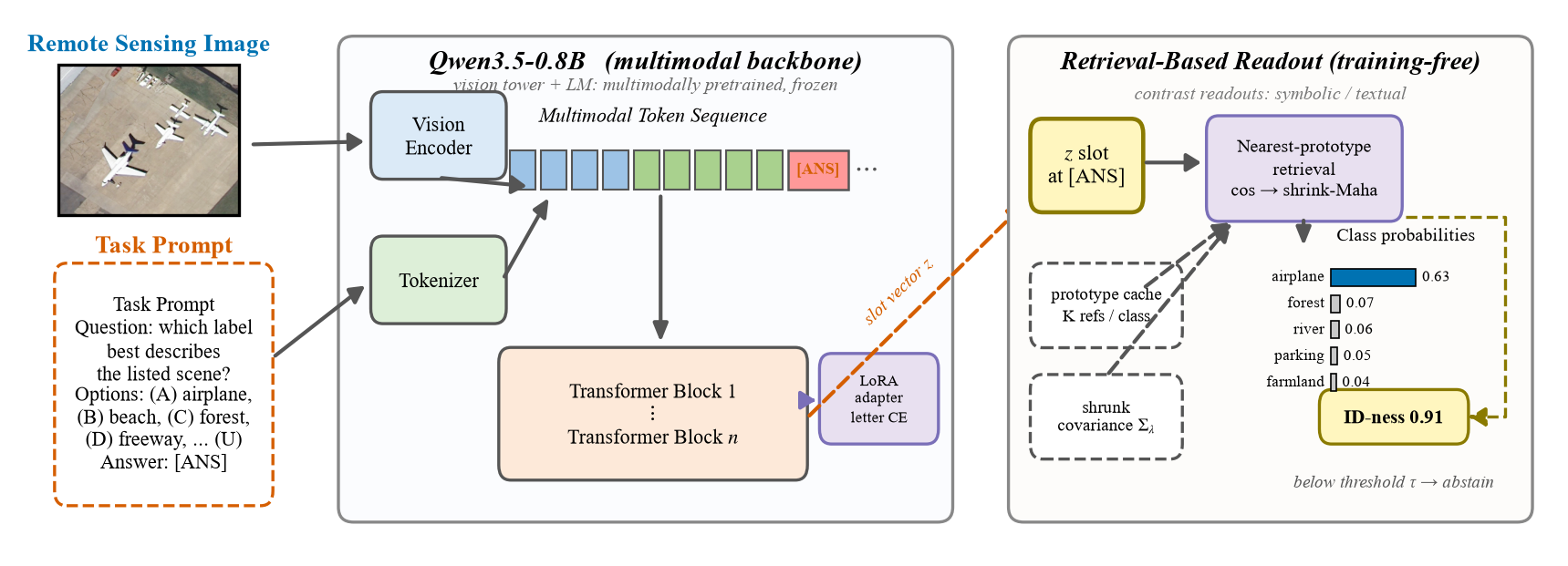}
  \caption{Pipeline overview: one shared forward pass produces the slot representation; the retrieval readout completes the decision over cached prototypes and provides the abstention channel.}
  \label{fig:pipeline}
\end{figure*}

\subsection{Preliminaries: One-Pass Decoding with Slot Binding}
\label{sec:prelim}

Let $C = \{c_1, \ldots, c_N\}$ be the candidate vocabulary. A one-pass decoder renders a prompt containing the image and an option list in which each class is bound to a letter token,
\begin{equation}
\begin{aligned}
\mathbf{p} ={}& \texttt{``Context \ldots\ Question: What is the scene?}\\
&\texttt{Options: (A) $c_1$ (B) $c_2$ \ldots\ Answer: [ANS]''}
\end{aligned}
\label{eq:prompt}
\end{equation}
where \texttt{[ANS]} is a registered special answer token appended after the option list. Driving the frozen MLLM conditioned on $\mathbf{p}$, the last-layer hidden state at the final position is taken as the slot representation
\begin{equation}
  \mathbf{z} = h_{L}(\texttt{[ANS]};\, \mathbf{p}) \in \mathbb{R}^{d},
  \label{eq:slot}
\end{equation}
which aggregates information from both the image and the option list via decoder-side self-attention. At classification time, all candidates are scored at once against the rows of the language-modeling head indexed by letter tokens,
\begin{equation}
  \hat{y} = \arg\max_{j}\; \bigl(W_{\mathrm{letter}_{j}}\, \mathbf{z}\bigr),
  \label{eq:score}
\end{equation}
where $\mathrm{letter}_{j}$ enumerates A, B, \ldots\ over a fixed alphabet. Slot-level fine-tuning minimizes cross-entropy on Eq.~\eqref{eq:score} over the training split with the base model frozen; a LoRA adapter suffices. Three properties follow directly from this construction: letter tokens carry no semantics of their own, class identity being injected entirely through the prompt context of Eq.~\eqref{eq:prompt}; the scoring space of Eq.~\eqref{eq:score} is exactly the letter set used in training, which makes it closed-set optimal yet devoid of any out-of-vocabulary mechanism; and the binding is coupled to the vocabulary---a model trained with a 30-class vocabulary facing a 21-class target vocabulary can only score via near-synonym overlap of class names, with no mechanism to rebind letters without retraining. Figure~\ref{fig:pipeline} summarizes the resulting pipeline: one shared forward pass produces the slot representation, and the retrieval readout and control readouts read the same cached representation at inference.

\paragraph{Data flow and pretraining.}
The backbone is Qwen3.5-0.8B (the series has no standalone technical report; its multimodal lineage is~\cite{qwenvl,qwen25vl,qwen3vl}), chosen for three reasons. First, the latency and memory budget of one-pass decision is directly determined by backbone scale; 0.8B is the smallest multimodal configuration of the series, measuring 232.49\,ms average per-image latency and 3.22\,GiB peak memory on an RTX 4090 (Section~\ref{sec:efficiency}), leaving headroom on the deployment side. Second, its vision tower and language model complete multimodal pretraining on large image--text corpora, supporting a registered \texttt{[ANS]} answer slot and LoRA~\cite{lora} adaptation out of the box. Third, it is of the same family and scale as the public configuration of the one-pass decider line~\cite{rsjev}, keeping our training pipeline and all comparisons on the same footing. Slot-level fine-tuning modifies no pretrained weight: the LoRA adapter is updated only on the training split with letter-token cross-entropy as the objective, and merged into the base after training. At inference, data flows as follows: the image is encoded by the vision tower into visual tokens; the prompt text (candidate list and \texttt{[ANS]} included) is encoded by the tokenizer into text tokens; the two sequences are concatenated and flow through the Transformer layers, producing the slot representation $\mathbf{z}$ at the last-layer hidden state of the \texttt{[ANS]} position; every decision thereafter---closed-set scoring, open-set ID-ness, candidate-set changes---reads only this one vector.

\subsection{The Retrieval Readout and Control Readouts}
\label{sec:readouts}

Our method, RSAJEV, is the retrieval-based prototype readout: it consumes the same $\mathbf{z}$ of Eq.~\eqref{eq:slot}, computing $\mathbf{z}$ is the only cost incurred, and the readout itself runs on cached $\mathbf{z}$ vectors. The symbolic and text readouts participate only as controls.

\paragraph{Retrieval-based prototype readout.}
For each class $c_j$, take $K$ reference images from the training split, forward them under the same option-rendered prompt as inference, and average and normalize their slot representations to obtain the class prototype
\begin{equation}
\begin{aligned}
\mathbf{p}_{j} &= \mathrm{normalize}\Bigl(\tfrac{1}{K}\textstyle\sum_{k} \mathbf{z}_{jk}\Bigr),\\
\hat{y} &= \arg\max_{j}\; \cos(\mathbf{z},\, \mathbf{p}_{j}).
\end{aligned}
\label{eq:proto}
\end{equation}
Classification is nearest-neighbor retrieval over the cached prototypes in the same representation space, isomorphic to retrieval-augmented generation (RAG) except that what is retrieved are class prototypes rather than documents, and the retrieval result directly yields a decision rather than text. To be clear, the prompt format stays as in Eq.~\eqref{eq:prompt}: candidate class names are still rendered as lettered options in the context, and the slot representation $\mathbf{z}$ is still conditioned on the full candidate list; what is replaced is only the scoring head reading $\mathbf{z}$---from trained letter-row scoring to retrieval over cached prototypes---not the prompt itself. Reference images come only from the training split, with a filename-level leakage check enforced in code; the reference cache is computed once. In our experiments, mean aggregation consistently outperforms medoid aggregation, which we attribute to the compactness of slot space; the $K$ sweep and the aggregation ablation are in Section~\ref{sec:ablations}.

\paragraph{Shrinkage-Mahalanobis readout.}
Cosine implies an isotropy assumption, whereas slot-level fine-tuning pulls the representation geometry toward the letter-readout direction, leaving a systematic deformation that can be whitened by a covariance estimated from the references; we therefore replace cosine with the Mahalanobis distance as the retrieval metric,
\begin{equation}
\begin{aligned}
\hat{y} &= \arg\max_{j}\; (\mathbf{z}-\mathbf{p}_{j})^{\top} \Sigma_{\lambda}^{-1} (\mathbf{z}-\mathbf{p}_{j}),\\
\Sigma_{\lambda} &= \Sigma + \tfrac{\lambda\,\mathrm{tr}(\Sigma)}{d}\, I,
\end{aligned}
\label{eq:maha}
\end{equation}
where $\Sigma$ is estimated from the slot representations of all references, involving no test statistics; the shrinkage term suppresses the ill-conditioned condition number of small-sample covariance estimation. $\lambda$ is chosen on held-out data, with a sensitivity analysis in Section~\ref{sec:ablations}.

\paragraph{Control readouts.}
The symbolic readout restricts the trained scoring of Eq.~\eqref{eq:score} to an arbitrary candidate subset $S \subseteq C$,
\begin{equation}
  \hat{y} = \arg\max_{c_{j}\in S}\; \bigl(W_{\mathrm{letter}_{j}}\, \mathbf{z}\bigr),
  \label{eq:restrict}
\end{equation}
where $\mathrm{letter}_{j}$ is the letter bound to $c_{j}$ in the prompt that produced $\mathbf{z}$. Restricting the argmax to the subset of rendered options realizes a new candidate set without re-rendering the prompt; Section~\ref{sec:opensetscore} will show that this pruning-based rescoring is not merely a convenience but measurably better than re-rendering. The text readout computes for each class a template sentence embedding (e.g., ``The remote sensing image shows \{name\}.''), forwarded as pure text through the same model and read out at the last token position; since both ends are last-layer hidden states, $\mathbf{z}$ and the class embedding $\mathbf{e}_{j}$ live in the same space and are compared by cosine similarity,
\begin{equation}
  \hat{y} = \arg\max_{j}\; \cos(\mathbf{z},\, \mathbf{e}_{j}).
  \label{eq:text}
\end{equation}
The embedding $\mathbf{e}_{j}$ is computed once per class---21 to 45 short-text forwards, image-independent; no letter binding is involved, so this readout applies to any vocabulary, including vocabularies never seen during slot training; subsequent tables denote it \cost{}.

The three readouts form a data--compute spectrum: the symbolic readout demands slot training on the target vocabulary, the text readout demands nothing but text, and the retrieval prototype readout demands $K$ references per class and no training at all.

\subsection{Open-Set Scoring and Prompt Robustness}
\label{sec:opensetscore}

Open-set rejection attaches an ID-ness score to every input. For the symbolic readout, natural candidates are the maximum softmax, the top-1--top-2 margin, and the negative entropy of the letter distribution; for the text and prototype readouts, the maximum cosine similarity of Eqs.~\eqref{eq:text} and~\eqref{eq:proto}. In deployment, ID-ness thresholds directly into a rejection rule: inputs below threshold $\tau$ output rejection instead of a class, with $\tau$ chosen on held-out in-domain data per the operating point's false-positive budget. Open-set quality is reported as AUROC and OSCR.

All these readouts share one representation, which raises the question of how much $\mathbf{z}$ depends on the prompt that produced it. The column-pruning experiment answers this without any extra forward pass: compare (i) $\mathbf{z}$ computed under the full $N$-option prompt, decoded via Eq.~\eqref{eq:restrict} restricted to subset $S$, with (ii) $\mathbf{z}$ recomputed under a prompt rendered with exactly $S$. The former changes only the readout; the latter changes the representation as well. In our 21-to-14-class reduction, pruning explains the entire accuracy gain of the symbolic readout while the representation change contributes negatively; for the text and prototype readouts, representation drift moves accuracy by at most 3.7 points. The slot representation is therefore prompt-robust, with a practical consequence uniform across all readouts: one forward pass under any fixed prompt suffices, and open-set rejection, vocabulary changes, and class additions are all realized as rescoring at zero inference cost.

\section{Experiments}
\label{sec:experiments}

\begin{figure*}[!t]
  \centering
  \includegraphics[width=\textwidth]{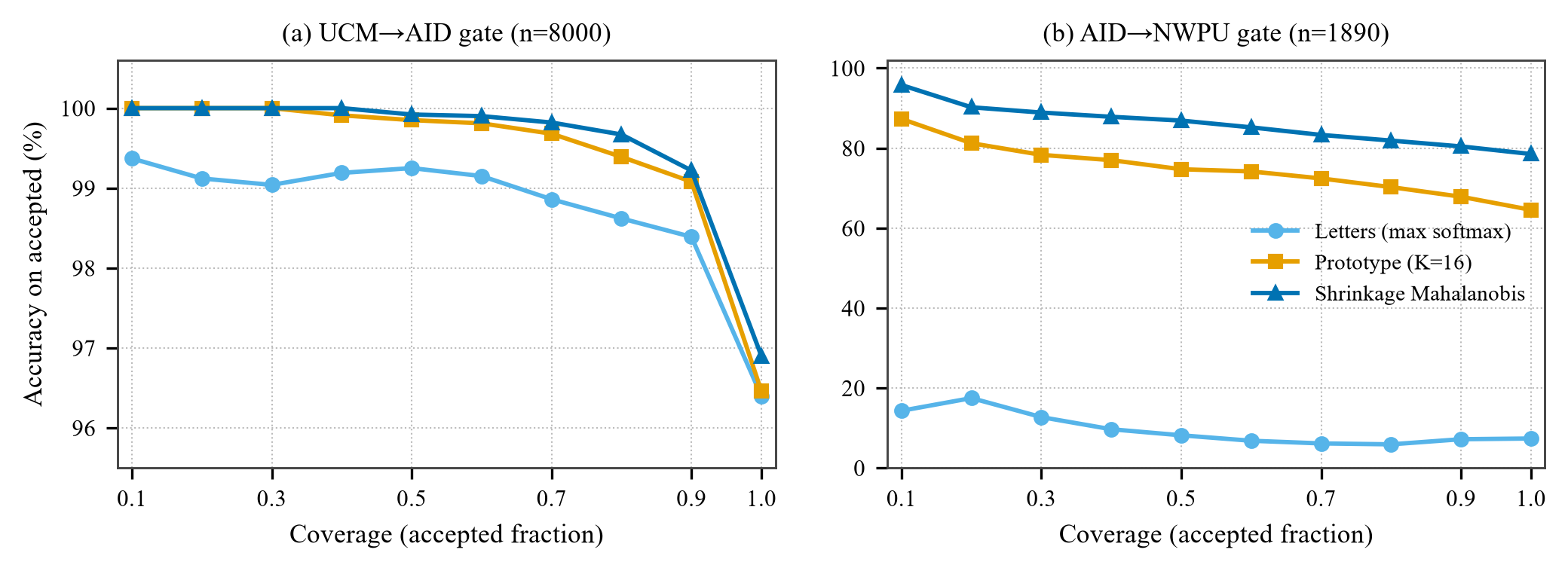}
  \caption{Coverage--accuracy curves: subset accuracy after accepting samples in descending order of ID-ness. (a) UCM$\to$AID gate; (b) AID$\to$NWPU gate---95.77 at 10\% coverage versus 87.30 for cosine and 14.29 for the symbolic head.}
  \label{fig:coverage}
\end{figure*}

\subsection{Datasets and Protocol}
\label{sec:datasets}

We evaluate on three standard remote sensing scene classification benchmarks: UC Merced Land Use~\cite{ucm} (UCM, 21 classes, 2{,}100 images, $256\times256$), Aerial Image Dataset~\cite{aid} (AID, 30 classes, 10{,}000 images, $600\times600$), and NWPU-RESISC45~\cite{nwpu} (NWPU, 45 classes, 31{,}500 images, $256\times256$). UCM uses a 50\%/50\% train/test split and AID a 20\%/80\% split, consistent with the official training protocol of~\cite{rsjev}; our splits are drawn with seed 42 and released with the code. NWPU serves only as an unseen target vocabulary. The prototype readout's reference set uses training-split images only, with a filename-level leakage assertion enforced in code; all control readouts (symbolic, text, prototype, Mahalanobis) run on exactly the same cached forward pass as our method; the official weights of~\cite{rsjev} are evaluated directly under their released configuration, and the UCM weights are trained under their official training protocol. Cross-dataset evaluation is zero adaptation on the target vocabulary, on the same footing as the officially released weights.

Open-set results are reported with three standard metrics: AUROC measures the threshold-free ranking quality of an ID-ness score placing seen before unseen---0.5 is the random line, and below 0.5 means inverted discrimination; FPR@95TPR is the fraction of unseen samples falsely accepted while retaining 95\% of seen samples; OSCR couples correct classification and correct rejection into a single aggregate score. Selective prediction is presented as coverage--accuracy curves (Fig.~\ref{fig:coverage}): samples are accepted in descending order of ID-ness and accuracy is reported on the accepted subset. $\lambda$ and $K$ are hyperparameters, not metrics; their protocols are given in Sections~\ref{sec:impl} and~\ref{sec:ablations}.

\subsection{Implementation Details}
\label{sec:impl}

All experiments are based on the officially released weights of the Qwen3.5-0.8B multimodal backbone. Slot fine-tuning uses a LoRA adapter trained with letter-token cross-entropy under the official training recipe (AdamW, 10 epochs); the adapter is merged into the base model before evaluation. All readouts are training-free and introduce no learnable parameters of their own. Every image is forwarded exactly once per candidate-set rendering, all readouts run on the cached slot representations, and readout comparison costs nothing beyond the shared forward pass. The prototype readout uses $K{=}16$ references per class with mean aggregation; the Mahalanobis variant uses a shared covariance estimated from the target-domain reference pool (16 per class---336 references for the UCM target), with the shrinkage coefficient $\lambda$ selected once on a shared validation grid by per-setting accuracy and no per-dataset tuning (ablation default $\lambda{=}0.02$; sensitivity in Section~\ref{sec:ablations}). Open-set evaluation uses UCM's 14-seen/7-unseen split; thresholds are calibrated only on held-out in-domain data; metric definitions are in Section~\ref{sec:datasets}.

\subsection{Closed-Set Benchmarks}
\label{sec:closedset}

The officially released weights measure 97.39\% OA and 97.16\% F1 on the AID test split, same-source-different-draw with the officially reported 96.56/96.30 and of consistent magnitude; 97.39 is the anchor of Table~\ref{tab:gate}'s in-domain row and of all AID in-domain measurements. We additionally train UCM weights as a second training source under the official training protocol, with training stopped by validation accuracy (optimum at 1 epoch); this checkpoint simultaneously serves as Table~\ref{tab:gate}'s UCM in-domain row and as the readout stress test under the weakest letter head. Table~\ref{tab:closedset} summarizes the closed-set numbers on both counts.

\begin{table*}[!t]
  \caption{Closed-Set Evaluation and Officially Released Numbers (\%). The Two AID Rows Use Different Test-Split Draws and Serve as Magnitude References}
  \label{tab:closedset}
  \centering
  \small
  \begin{tabular}{lcccc}
    \toprule
    Setting & UCM OA$\uparrow$ & UCM F1$\uparrow$ & AID OA$\uparrow$ & AID F1$\uparrow$ \\
    \midrule
    Official report~\cite{rsjev}      & 97.71 & 97.72 & 96.56          & 96.30 \\
    Ours: official-weights evaluation & --    & --    & \textbf{97.39} & \textbf{97.16} \\
    \bottomrule
  \end{tabular}
\end{table*}

\subsection{The Cross-Dataset Decision Gate}
\label{sec:gate}

Table~\ref{tab:gate} freezes one forward pass per target image and swaps only the readout. Rows 1--4 are cross-dataset (the target vocabulary never appeared in slot training); the last two rows are in-domain---the UCM row's weights and references both come from the UCM training set, and the AID row uses the official AID weights with AID training references (its letter-head column is the 97.39 of Section~\ref{sec:closedset}, same split and prompt); the two rows are not directly comparable to the cross-dataset rows.

\paragraph{Direction design.}
The gate is the complete matrix of two training sources (the officially released AID checkpoint and our UCM checkpoint trained under the same official protocol) crossed with two purely unseen target vocabularies (UCM and NWPU), plus two in-domain rows (UCM$\to$UCM and AID$\to$AID) that separate the contribution of the reference distribution and provide controls on the trained readout's home turf. The AID directions' zero-adaptation numbers are strictly comparable to published figures; the UCM directions repeat the same measurement from another training source to confirm that the conclusions do not depend on a particular source. The two target vocabularies span 21 and 45 classes, with NWPU the hardest unseen target by class count and intra-class variance. NWPU never participates in any training, so we do not report its in-domain accuracy: the Mahalanobis readouts of the two NWPU rows in Table~\ref{tab:gate} (78.52 and 77.99) are the OA on the NWPU vocabulary, measuring precisely decision quality ``facing a 45-class world never trained on''---the rows that best exhibit the deployment value of the training-free readout.

\begin{table*}[!t]
  \caption{Cross-Dataset Decision Gate: One Frozen Forward Pass and Reference Cache, Swapping Only the Scoring Head, Comparing Four Readouts Row by Row (OA, \%; $\lambda$ Selected on the Shared Validation Grid)}
  \label{tab:gate}
  \centering
  \small
  \begin{tabular}{lccccc}
    \toprule
    Direction & letters & \cost{} & \proto{} & \maha{} & $\Delta$ \\
    \midrule
    AID $\to$ UCM (cross, 21 classes)    & 21.52 & 57.33 & 78.76 & \textbf{89.43} & $+10.7$ \\
    UCM $\to$ AID (cross, 30 classes)    & 96.40 & 91.55 & 96.46 & \textbf{96.90} & $+0.4$  \\
    AID $\to$ NWPU (cross, 45 classes)   & 7.35  & 39.47 & 64.55 & \textbf{78.52} & $+14.0$ \\
    UCM $\to$ NWPU (cross, 45 classes)   & 34.29 & 22.80 & 64.50 & \textbf{77.99} & $+13.5$ \\
    \midrule
    UCM $\to$ UCM (in-domain)            & 78.10 & 38.10 & 82.86 & \textbf{89.05} & $+6.2$  \\
    AID $\to$ AID (in-domain)            & 97.39 & 96.81 & \textbf{97.51} & 97.40 & $+0.1$  \\
    \bottomrule
  \end{tabular}
\end{table*}

\begin{figure*}[!t]
  \centering
  \includegraphics[width=\textwidth]{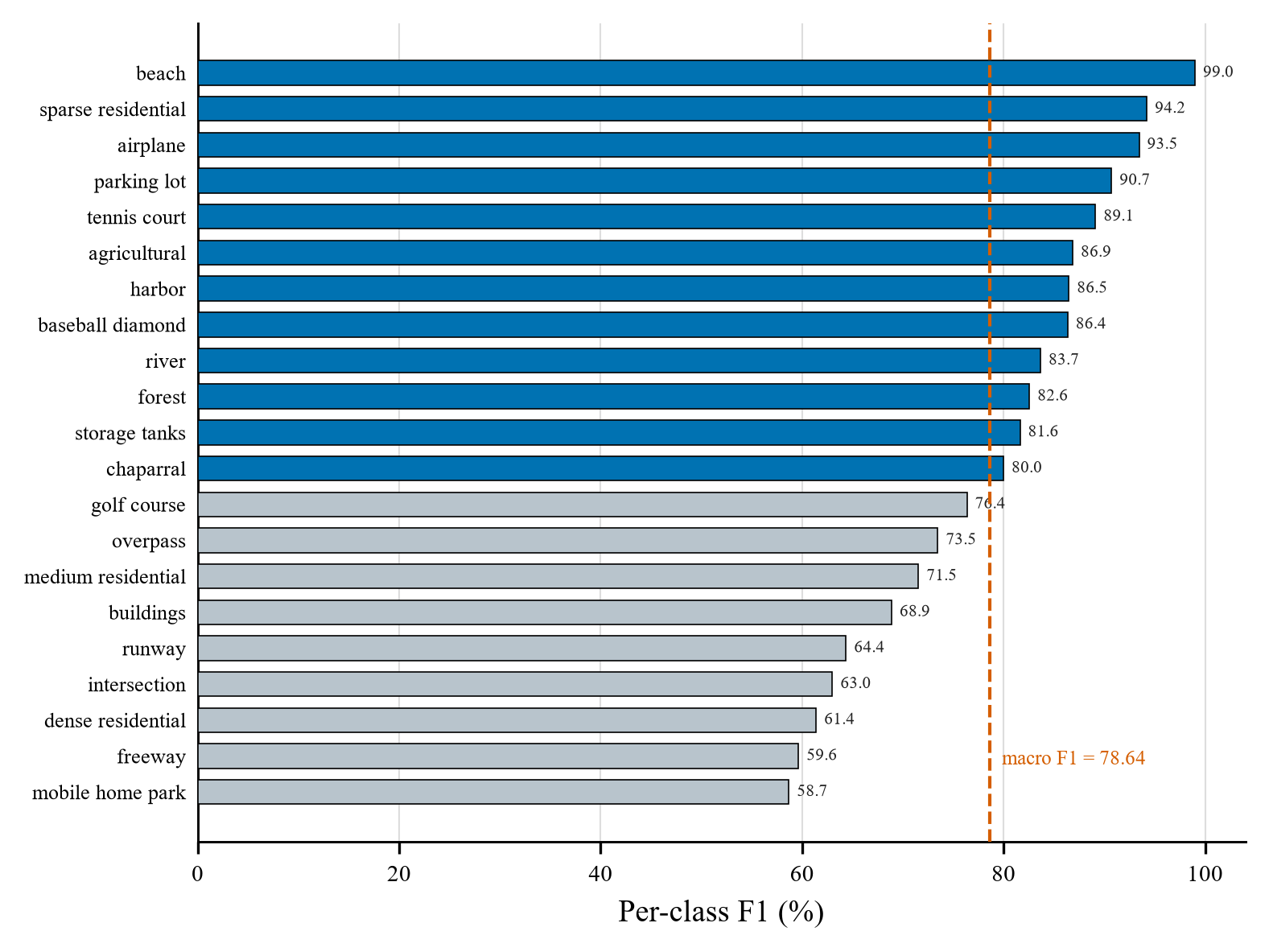}
  \caption{Per-Class F1 of the Prototype Readout ($K{=}16$) on the UCM Cross-Dataset Gate (Derived From the Confusion Matrix): 13/21 Classes Exceed the 78.64 Macro-Average}
  \label{fig:perclass}
\end{figure*}

Two regularities support this section. The symbolic readout fails in every cross-dataset direction (21.52 for AID$\to$UCM, 96.40 for UCM$\to$AID, 7.35 for AID$\to$NWPU, 34.29 for UCM$\to$NWPU; chance is 2.2--4.8), its fate decided by training trajectory rather than vocabulary distance: our UCM weights trained under the same official protocol read the unseen 30-class AID vocabulary at 96.40, while the official AID weights collapse to 21.52 on UCM and 7.35 on NWPU. A readout whose zero-adaptation behavior spans 89 points across shipped weights cannot be trusted outside its training label space; the retrieval readout never queries the letter head, its range across all six rows being only 19.4 points. The text readout, though also training-free, swings 74 points across target domains (22.80--96.81), an instability that reappears in the open-set evaluation of Section~\ref{sec:openset}---training-free does not automatically mean stable; stability comes from the concrete retrieval target of a visual reference cache. Prototype and Mahalanobis readouts take the top two spots in every row, the Mahalanobis correction beating the cosine prototype ($+10.7$, $+0.4$, $+14.0$, and $+13.5$ across the four cross-dataset directions; $+6.2$ in-domain on UCM), placing second by 0.11 points on the nearly isotropic AID in-domain---harmless---with an even more lopsided advantage over the also-training-free text readout ($+5.4$ to $+55.2$); pure geometric correction narrows the gap to fully supervised accuracy on UCM from 19.0 to 8.3 points. On the 30-class AID target the Mahalanobis readout reaches 96.90, the best zero-adaptation number in the table, with an almost flat $\lambda$ curve (96.83--96.94) indicating near-isotropy of slot space at this scale. The in-domain rows add four observations: this checkpoint's own letter head is only 78.10, overtaken by the training-free prototype (82.86) and Mahalanobis (89.05) readouts by 4.8 and 11.0 points---even within the training distribution, the retrieval readout beats the trained letter head; when weights and references share the training distribution, visual prototypes gain another 4.1 points over the cross-dataset rows; the text readout drops to 38.10, showing that slot fine-tuning pulls the head toward letters at the cost of text-aligned geometry---the two training-free readouts are not interchangeable, visual references being strictly stronger; and the Mahalanobis readout transfers as-is, 89.05 in-domain versus 89.43 cross-dataset, 6.2 points above the in-domain cosine prototype, confirming that anisotropy correction is a property of slot space rather than of any particular weight--reference pairing. The AID in-domain row gives the other side of the same conclusion on the trained readout's home turf: the retrieval prototype readout's 97.51 exceeds the letter head's 97.39---even with official weights under the official protocol, training-free retrieval with 16 references per class does not lose to the trained readout; the Mahalanobis $\lambda$ curve is nearly flat (97.19--97.50, $\lambda{=}1.0$ optimal), confirming with the UCM$\to$AID gate that AID slot space is nearly isotropic; the text readout's 96.81 is within 0.6 of the letter head, in contrast to the in-domain UCM row's 38.10. Score fusion never beats the best single component (79.43 at the validation-selected $\alpha{=}0.25$, below the best single retrieval readout's 89.43), consistent with Section~\ref{sec:ablations}. Figure~\ref{fig:perclass} aggregates the gate's confusion matrix into per-class F1: 13/21 classes exceed the 78.64 macro-average, with the tail (mobile home park 58.7, freeway 59.6, dense residential 61.4) concentrated in morphologically similar residential and traffic-corridor classes.

\subsection{Open-Set Recognition}
\label{sec:openset}

\begin{figure*}[!t]
  \centering
  \includegraphics[width=\textwidth]{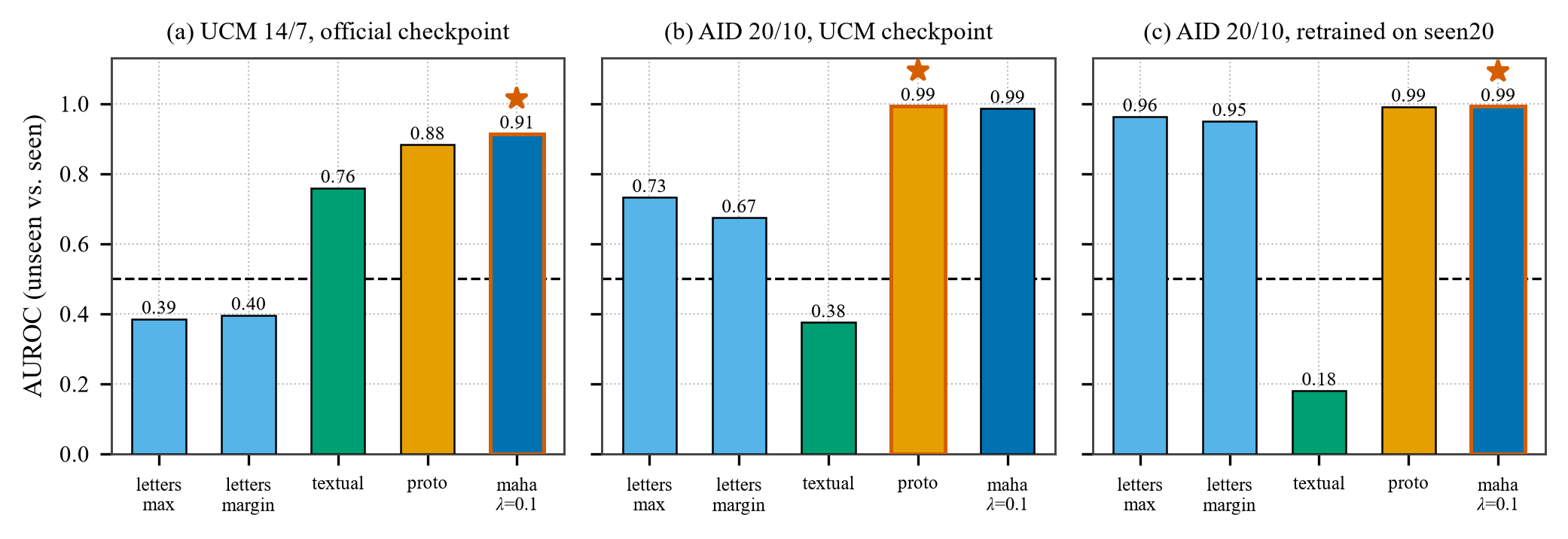}
  \caption{AUROC of Five ID-ness Scorers Under Three Open-Set Settings: UCM 14/7, AID 20/10 Cross-Weights, AID 20/10 Clean Weights}
  \label{fig:auroc}
\end{figure*}

\begin{table}[!t]
  \caption{UCM 14 Seen/7 Unseen, Official Weights; Out-of-Vocabulary Weight Setting (Deployed Weights Whose Training Vocabulary Excludes the Target Labels)}
  \label{tab:ucmoov}
  \centering
  \footnotesize
  \setlength{\tabcolsep}{4pt}
  \begin{tabular}{lccc}
    \toprule
    ID-ness score & AUROC$\uparrow$ & FPR@95TPR$\downarrow$ & OSCR$\uparrow$ \\
    \midrule
    letters (max softmax)   & 0.3853         & 1.0000         & 0.1291 \\
    \cost{} (text control)  & 0.7593         & 0.6214         & 0.2304 \\
    \proto{} (cosine proto) & 0.8826         & 0.6029         & 0.2661 \\
    \maha{} (ours, $\lambda{=}0.05$) & \textbf{0.9147} & \textbf{0.4671} & \textbf{0.2680} \\
    \bottomrule
  \end{tabular}
\end{table}

Symbolic confidence separates seen from unseen in the wrong direction: the letter head of discriminative training is negatively correlated with novelty. The failure cannot be repaired by threshold choice, because the letter head's outcome space never contains ``none of the above''; the symbolic row of Table~\ref{tab:ucmoov} is the measurable shape of silent hallucination. Prototype similarity degrades gracefully, and the Mahalanobis readout adds another 3.2 points of AUROC over cosine, cutting FPR at 95\% TPR by 13.6 points. OSCR couples rejection with closed-set correctness and, in this setting, is capped by the letter head's 27.4\% seen-class closed-set accuracy---all four scorers' OSCR falls below 0.27---pinning the bottleneck precisely on the classification side rather than the rejection side: Mahalanobis scoring has already pushed seen/unseen separation to 0.9147 AUROC, and once the classification side switches to the retrieval readout on the same representation (Table~\ref{tab:gate}: 89.43\%), the abstention interface becomes usable again. This is exactly why we replace the readout rather than patch thresholds.

\begin{table}[!t]
  \caption{UCM 14 Seen/7 Unseen, Weights Retrained on the 14 Seen Classes Only (2 Epochs, Never Seeing the 7 OOD Classes); Clean-Denominator Setting, Testing Whether the Conclusions of Table~\ref{tab:ucmoov} Hold}
  \label{tab:ucmclean}
  \centering
  \footnotesize
  \setlength{\tabcolsep}{4pt}
  \begin{tabular}{lccc}
    \toprule
    ID-ness score & AUROC$\uparrow$ & FPR@95TPR$\downarrow$ & OSCR$\uparrow$ \\
    \midrule
    letters (max softmax)   & 0.7642         & 0.6729         & 0.7151 \\
    \cost{} (text control)  & 0.6545         & 0.6486         & 0.5394 \\
    \proto{} (cosine proto) & 0.9121         & 0.3814         & 0.8123 \\
    \maha{} (ours, $\lambda{=}0.1$) & \textbf{0.9426} & \textbf{0.2900} & \textbf{0.8304} \\
    \bottomrule
  \end{tabular}
\end{table}

The clean-denominator setting, with a model trained only on the 14 seen classes, confirms the cross-weight finding: the symbolic head now behaves normally (AUROC 0.76; the inversion is a pseudo-open-set artifact of out-of-vocabulary weights), while prototype and Mahalanobis readouts remain the strongest ID-ness scores, the geometric correction transferring again (0.9426, 3.0 above cosine, FPR down 9.1 points), and the OSCR ranking consistent (Mahalanobis 0.8304, prototype 0.8123, letters 0.7151). Score fusion again fails to beat the best single component (0.51).

\begin{table}[!t]
  \caption{AID 20 Seen/10 Unseen (Seed-42 Split, 20-Option Prompts, $n{=}8000$), UCM-Trained Weights Never Having Seen AID Classes; Out-of-Vocabulary Weight Setting}
  \label{tab:aidoov}
  \centering
  \footnotesize
  \setlength{\tabcolsep}{4pt}
  \begin{tabular}{lccc}
    \toprule
    ID-ness score & AUROC$\uparrow$ & FPR@95TPR$\downarrow$ & OSCR$\uparrow$ \\
    \midrule
    letters (max softmax)   & 0.7323                 & 0.9208                 & 0.5755 \\
    \cost{} (text control)  & 0.3750                 & 0.9990                 & 0.2897 \\
    \proto{} (cosine proto) & \textbf{0.9936}        & \textbf{0.0219}        & \textbf{0.7524} \\
    \maha{} (ours, $\lambda{=}0.1$) & 0.9865         & 0.0627                 & 0.7486 \\
    \bottomrule
  \end{tabular}
\end{table}

\begin{table}[!t]
  \caption{AID 20 Seen/10 Unseen, Weights Retrained on the 20 Seen Classes Only (2 Epochs, Never Seeing the 10 OOD Classes); Clean-Denominator Setting}
  \label{tab:aidclean}
  \centering
  \footnotesize
  \setlength{\tabcolsep}{4pt}
  \begin{tabular}{lccc}
    \toprule
    ID-ness score & AUROC$\uparrow$ & FPR@95TPR$\downarrow$ & OSCR$\uparrow$ \\
    \midrule
    letters (max softmax)   & 0.9630         & 0.1176                 & 0.9444 \\
    \cost{} (text control)  & 0.1806         & 1.0000                 & 0.1763 \\
    \proto{} (cosine proto) & 0.9911         & 0.0318                 & 0.9699 \\
    \maha{} (ours, $\lambda{=}0.5$) & \textbf{0.9937} & \textbf{0.0249} & \textbf{0.9715} \\
    \bottomrule
  \end{tabular}
\end{table}

Two symmetric tables complete the open-set picture on the second dataset. Under UCM-trained weights (Table~\ref{tab:aidoov}), the symbolic head behaves normally---it has never seen any AID class---but remains weak (0.73, FPR 0.92), while prototype matching reaches 0.9936 with FPR at 95\% TPR of 2.2\%: AID's unseen classes are nearly perfectly separated in slot space. The Mahalanobis correction fails to beat cosine here (0.9865 at $\lambda{=}0.1$); consistent with the nearly flat $\lambda$ curve of the UCM$\to$AID gate (96.83--96.94), the AID target's slot space is close to isotropic and already near the ceiling, leaving whitening nothing to correct: the geometric prior helps where anisotropy exists and is harmless where it does not. The text score fails outright as an ID-ness signal on this target (0.3750 and 0.1806, the latter inverted), a degradation not observed on UCM; template cosine happens to be dataset-sensitive exactly where prototype matching is strongest. Under clean retrained weights (Table~\ref{tab:aidclean}), the symbolic head is again normal and much stronger (0.9630, FPR 11.8\%, seen-class closed-set accuracy 0.9753), and the Mahalanobis correction regains a small advantage with stronger shrinkage (0.9937 at $\lambda{=}0.5$ versus cosine's 0.9911, FPR down from 3.2\% to 2.5\%). Score fusion fails again (0.47 and 0.49, neither beating the best single component in the corresponding table). The ranking pattern of Tables~\ref{tab:ucmoov}--\ref{tab:aidoov}---retrieval readouts first, symbolic unreliable, fusion harmful---holds on AID, with OSCR ranking consistent with AUROC (Table~\ref{tab:aidoov}'s prototype first at 0.7524, Table~\ref{tab:aidclean}'s Mahalanobis first at 0.9715), plus one deployment-friendly observation: near the 0.99 ceiling the Mahalanobis-over-cosine margin narrows to noise level---the two retrieval readouts simultaneously approach the representation's separation ceiling, and within this range the choice between readouts is no longer sensitive, the hyperparameter burden approaching zero. Figure~\ref{fig:auroc} presents three of the four open-set settings side by side (the UCM clean-denominator setting of Table~\ref{tab:ucmclean} shares the construction of Fig.~\ref{fig:auroc}(a) with a consistent ranking and is not plotted again), with five ID-ness scorers per setting.

\subsection{Ablations}
\label{sec:ablations}

\begin{figure*}[!t]
  \centering
  \includegraphics[width=\textwidth]{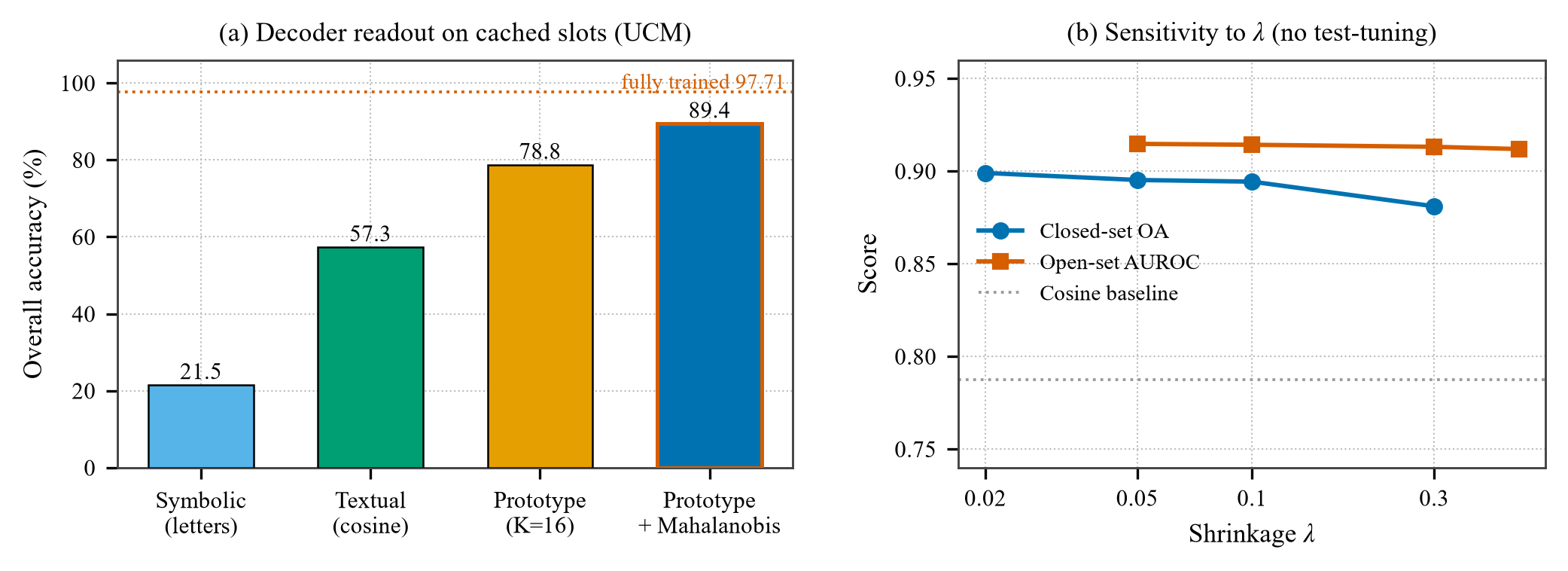}
  \caption{Readout Ablation on the Shared Representation (a) and Shrinkage-Mahalanobis $\lambda$ Sensitivity (b)}
  \label{fig:ablation}
\end{figure*}

\begin{figure}[!t]
  \centering
  \includegraphics[width=\linewidth]{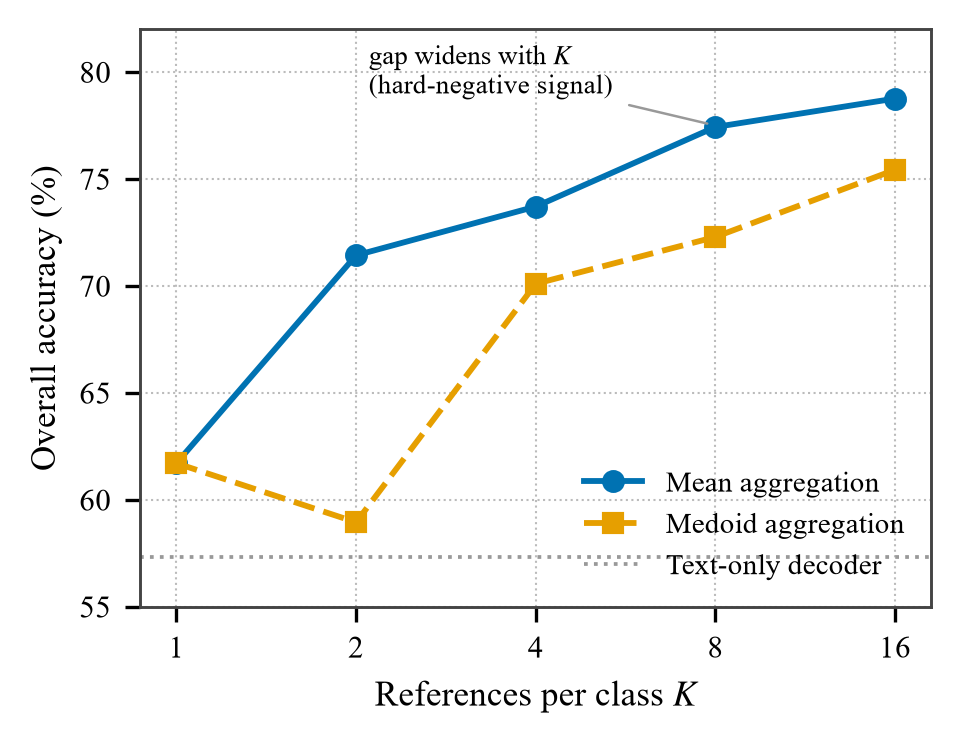}
  \caption{Prototype-Readout Accuracy Versus the Number of References $K$: Mean and Medoid Aggregation (UCM Gate)}
  \label{fig:ksweep}
\end{figure}

The ablations span six dimensions, all on the shared representation and introducing no new forwards: readout form (Table~\ref{tab:ablation} and Fig.~\ref{fig:ablation}(a)), the number of references $K$ (Fig.~\ref{fig:ksweep}), aggregation (mean vs.\ medoid, Fig.~\ref{fig:ksweep}), the shrinkage coefficient $\lambda$ (Fig.~\ref{fig:ablation}(b)), reference content (typicality filtering, this section), and readout combination (score fusion, this section).

The default readout chooses shrinkage Mahalanobis over cosine on the basis of two independently testable priors, argued in this section in the order ``argue, measure, adjudicate.'' Geometrically, slot-level fine-tuning takes letter-token cross-entropy as its sole objective, systematically pulling the last-layer geometry toward the trained letter-readout direction and leaving an anisotropic deformation: if this argument holds, whitening with a reference-estimated covariance (involving no test statistics) should bring large and robust gains whose size grows and shrinks with the target's anisotropy. Statistically, estimating a $d$-dimensional covariance from 16 references per class and a pool of several hundred samples makes direct inversion necessarily ill-conditioned; the shrinkage term of Eq.~\eqref{eq:maha} regularizes the spectrum by $\lambda\,\mathrm{tr}(\Sigma)/d\, I$---a standard correction in high-dimensional small-sample settings. The two priors jointly yield a falsifiable prediction: the Mahalanobis-over-cosine gain should be large and robust, vary across targets, and not depend on fine tuning.

\begin{table}[!t]
  \caption{Readout Ablation on UCM Closed-Set With 16 References Per Class (OA, \%). Geometric Correction Contributes About $+11$ Points; Every Cache-Construction and Score-Fusion Variant Falls Short of the Naive Prototype---the Gain Comes From Metric Geometry, Not Cache Content}
  \label{tab:ablation}
  \centering
  \footnotesize
  \setlength{\tabcolsep}{4pt}
  \begin{tabular}{lc}
    \toprule
    Readout variant & OA$\uparrow$ \\
    \midrule
    Cosine prototype (= L2 on normalized slots) & 78.76 \\
    $k$-means multi-prototype ($m{=}5$)         & 80.10 \\
    Typicality-filtered references (top-12)     & 78.00 \\
    Prototype + cosine score fusion             & 77.71 \\
    Shrinkage Mahalanobis ($\lambda{=}0.1$)     & 89.43 \\
    Shrinkage Mahalanobis ($\lambda{=}0.02$, default) & \textbf{89.90} \\
    \bottomrule
  \end{tabular}
\end{table}

Measurement confirms the predictions item by item. Whitening lifts UCM from 78.76 to 89.43--89.90, robust across the entire $\lambda$ grid (88.1--89.9, Fig.~\ref{fig:ablation}(b))---a property of the representation rather than a tunable advantage; the gain's magnitude exactly tracks anisotropy (Table~\ref{tab:gate}: $+10.7$ on UCM, $+14.0$ and $+13.5$ on NWPU, $+6.2$ in-domain, only $+0.4$ on the nearly isotropic AID), vanishing as predicted on targets without deformation. Typicality-based reference filtering (78.00 at top-12) and prototype-plus-cosine score fusion (77.71, below its own cosine component's 78.76) both fail to beat the naive prototype readout, confirming that the constraint sits in the readout rather than the cache content; $k$-means multi-prototype ($m{=}5$) pushes 78.76 only to 80.10, far short of the whitened 89.43--89.90, showing that the gain comes from metric geometry rather than cache construction. The same holds on the open-set cache: naive $z$-score fusion of cosine and Mahalanobis scores falls to 0.5423 AUROC while Mahalanobis alone reaches 0.9147; cosine and prototype matching overlap informationally in the same vector subspace, the Mahalanobis correction being a replacement for---not a complement to---the cosine readout. Aggregating all ten closed- and open-set settings (Tables~\ref{tab:gate}--\ref{tab:ablation}), shrinkage Mahalanobis ranks first eight times and second twice (by 0.0071 AUROC in Table~\ref{tab:aidoov} and 0.11 points of OA in the AID in-domain row), never leaving the top two: on this basis this paper takes shrinkage Mahalanobis as the default readout.

The same evidence is re-examined through the selective-prediction lens (Fig.~\ref{fig:coverage}): samples are accepted from highest to lowest ID-ness per scorer, and accuracy is reported on the accepted subset. On the UCM$\to$AID gate ($n{=}8000$) all three readouts hug the ceiling---100\% for prototype and Mahalanobis at 10\% coverage, 99.37\% for the symbolic head; separation unfolds on the harder target---on the AID$\to$NWPU gate the symbolic head's score is nearly unrelated to correctness (7.35 at full coverage, 14.29 at best in the highest-coverage segment), the cosine prototype lifts 10\%-coverage accuracy to 87.30, and shrinkage Mahalanobis further to 95.77 (86.88 versus 74.71 at 50\% coverage). The ranking quality of ID-ness is of the same origin as readout accuracy: geometric correction improves both the decision and the ranking of ``which samples to abstain on,'' so the abstention channel needs no additional uncertainty estimator.

\subsection{Efficiency}
\label{sec:efficiency}

All readouts share one forward pass per image, so efficiency is a property of the representation pipeline, not of any readout. Measured on an RTX 4090 (batch size 1, full fp32 pipeline: image preprocessing, prompt rendering, vision tower, language model, and slot readout; 30 UCM test images after 5 warm-up runs), the pipeline averages 232.49\,ms per image (minimum 184.71, median 233.46\,ms), 4.3 FPS, with 3.22\,GiB peak memory (3.31\,GiB reserved). The published comparison figure comes from an A40 with looser accounting (template caching, no re-rendering); this paper uniformly adopts the full fp32 pipeline including prompt re-rendering, the two not directly compared---232.49\,ms is thus a conservative upper bound. The cached slot representations make every readout-level operation (open-set rejection, vocabulary changes, class additions, Mahalanobis rescoring) an offline matrix computation at zero marginal inference cost; throughput is a ceiling, not a target: prompt caching, half-precision inference, and shorter option lists all directly lower the 232\,ms figure without touching the readout.

\section{Conclusion}
\label{sec:conclusion}

We formulate one-pass MLLM scene classifiers as a shared slot representation plus a retrieval-based training-free readout---instantiated as RSAJEV---and measure the two components separately under closed-set, cross-dataset, and open-set protocols. The representation is an asset: it is prompt-robust, and when the letter head swings from 78.1 at the weakest checkpoint to 97.7 at the official weights, the training-free Mahalanobis readout on the weakest-letter-head checkpoint still reaches 89.05, built from cached references alone; on official AID in-domain, where the trained readout is at its strongest, the retrieval prototype readout (97.51) still exceeds the letter head (97.39). The symbolic readout is a liability: its zero-adaptation accuracy spans 89 points across four cross-dataset directions with the training trajectory (7.35--96.40); its confidence is unusable for open-set rejection, negatively correlated with novelty under out-of-vocabulary weights (AUROC 0.38) and still below prototype similarity under clean weights (UCM 0.76 vs.\ 0.91, AID 0.96 vs.\ 0.99). In both cases the interface cannot abstain---the discriminative form of the silent hallucination plaguing generative MLLMs; the retrieval readout removes this failure at the mechanism level rather than patching it, because its ID-ness score is computed from the representation itself, not emitted by a trained head. Swapping the readout recovers the gap without a single gradient step: 16 references per class lift UCM cross-vocabulary accuracy from 21.52 to 89.43, the shrinkage-Mahalanobis gain over the cosine prototype scales with anisotropy ($+0.4$ to $+14.0$ points) and is harmless where anisotropy is absent, opens a 55.2-point lead over the text control readout on the most imbalanced direction, and open-set AUROC reaches 0.91--0.99 on both datasets with the retrieval readout first in all four open-set settings; the coverage--accuracy curves show that this ranking advantage holds across all abstention operating points (Fig.~\ref{fig:coverage}). Because each readout consumes only one cached forward pass per image (232.49\,ms average, 3.22\,GiB peak on an RTX 4090), open-set rejection, vocabulary changes, and class additions all reduce to offline matrix computation at zero marginal inference cost. The method's entire data overhead is $K$ reference images per class and a reference pool in the hundreds---versus the full training split and multi-round gradient descent required by the symbolic readout, an order-of-magnitude smaller deployment cost whose floor is set only by the sample size supporting covariance estimation; the open-set margin narrows near the separation ceiling, indicating that the retrieval readout has approached this representation's discriminative limit, with the remaining headroom on the representation side rather than the readout side---a clear division of labor for future work. Extending slot readout to dense prediction tasks and multi-label aerial vocabularies is the natural next step.

\bibliographystyle{IEEEtran}
\bibliography{refs}

@misc{gpt4,
  author = {{OpenAI}},
  title  = {{GPT-4} Technical Report},
  year   = {2023},
  note   = {arXiv:2303.08774, doi:10.48550/arXiv.2303.08774}
}

@inproceedings{flamingo,
  author    = {Jean-Baptiste Alayrac and others},
  title     = {Flamingo: A Visual Language Model for Few-Shot Learning},
  booktitle = {Proc. NeurIPS},
  year      = {2022},
  pages     = {23716--23736},
  note      = {doi:10.52202/068431-1723}
}

@inproceedings{llava,
  author    = {Haotian Liu and Chunyuan Li and Qingyang Wu and Yong Jae Lee},
  title     = {Visual Instruction Tuning},
  booktitle = {Proc. NeurIPS},
  year      = {2023},
  note      = {doi:10.52202/075280-1516}
}

@inproceedings{geochat,
  author    = {Kartik Kuckreja and Muhammad Sohail Danish and M. Naseer and S. Das and Salman Khan and Fahad Shahbaz Khan},
  title     = {{GeoChat}: Grounded Large Vision-Language Model for Remote Sensing},
  booktitle = {Proc. IEEE/CVF CVPR},
  year      = {2024},
  pages     = {27831--27840},
  note      = {doi:10.1109/CVPR52733.2024.02629}
}

@inproceedings{lhrsbot,
  author    = {Dilxat Muhtar and Zhenshi Li and Feng Gu and Xueliang Zhang and Pengxin Xiao},
  title     = {{LHRS-Bot}: Empowering Remote Sensing with {VGI}-Enhanced Large Multimodal Language Model},
  booktitle = {Proc. ECCV},
  year      = {2024},
  pages     = {440--457},
  note      = {doi:10.1007/978-3-031-72904-1\_26}
}

@misc{rsjev,
  author = {D. Si and D. Wang and M. Xu and J. Zhang and B. Du and L. Zhang},
  title  = {{RSJEV}: Discriminative Remote Sensing Scene Classification with Multimodal Large Language Models},
  year   = {2026},
  note   = {arXiv:2610.08539, doi:10.48550/arXiv.2610.08539}
}

@article{hallsurvey,
  author  = {Lei Huang and others},
  title   = {A Survey on Hallucination in Large Language Models: Principles, Taxonomy, Challenges, and Open Questions},
  journal = {ACM Trans. Inf. Syst.},
  volume  = {43},
  number  = {2},
  year    = {2025},
  note    = {doi:10.1145/3703155}
}

@inproceedings{imagenet,
  author    = {Jia Deng and Wei Dong and Richard Socher and Li-Jia Li and Kai Li and Li Fei-Fei},
  title     = {{ImageNet}: A Large-Scale Hierarchical Image Database},
  booktitle = {Proc. IEEE CVPR},
  year      = {2009},
  pages     = {248--255},
  note      = {doi:10.1109/CVPR.2009.5206848}
}

@article{alexnet,
  author  = {Alex Krizhevsky and Ilya Sutskever and Geoffrey E. Hinton},
  title   = {{ImageNet} Classification with Deep Convolutional Neural Networks},
  journal = {Commun. ACM},
  volume  = {60},
  number  = {6},
  pages   = {84--90},
  year    = {2017},
  note    = {doi:10.1145/3065386}
}

@inproceedings{vgg,
  author    = {Karen Simonyan and Andrew Zisserman},
  title     = {Very Deep Convolutional Networks for Large-Scale Image Recognition},
  booktitle = {Proc. ICLR},
  year      = {2015},
  note      = {doi:10.48550/arXiv.1409.1556}
}

@inproceedings{resnet,
  author    = {Kaiming He and Xiangyu Zhang and Shaoqing Ren and Jian Sun},
  title     = {Deep Residual Learning for Image Recognition},
  booktitle = {Proc. IEEE/CVF CVPR},
  year      = {2016},
  pages     = {770--778},
  note      = {doi:10.1109/CVPR.2016.90}
}

@inproceedings{vit,
  author    = {Alexey Dosovitskiy and others},
  title     = {An Image is Worth 16x16 Words: Transformers for Image Recognition at Scale},
  booktitle = {Proc. ICLR},
  year      = {2021},
  note      = {doi:10.48550/arXiv.2010.11929}
}

@inproceedings{swin,
  author    = {Ze Liu and Yutong Lin and Yue Cao and Han Hu and Yixuan Zhang and Zheng Li and Wei Liu},
  title     = {Swin Transformer: Hierarchical Vision Transformer Using Shifted Windows},
  booktitle = {Proc. IEEE/CVF ICCV},
  year      = {2021},
  pages     = {10012--10022},
  note      = {doi:10.1109/ICCV48922.2021.00986}
}

@inproceedings{dino,
  author    = {Mathilde Caron and Hugo Touvron and Ishan Misra and Herv\'e J\'egou and Julien Mairal and Piotr Bojanowski and Armand Joulin},
  title     = {Emerging Properties in Self-Supervised Vision Transformers},
  booktitle = {Proc. IEEE/CVF ICCV},
  year      = {2021},
  pages     = {9650--9660},
  note      = {doi:10.1109/ICCV48922.2021.00951}
}

@inproceedings{mae,
  author    = {Kaiming He and Xinlei Chen and Saining Xie and Yanghao Li and Piotr Doll\'ar and Ross Girshick},
  title     = {Masked Autoencoders Are Scalable Vision Learners},
  booktitle = {Proc. IEEE/CVF CVPR},
  year      = {2022},
  pages     = {16000--16009},
  note      = {doi:10.1109/CVPR52688.2022.01553}
}

@inproceedings{clip,
  author    = {Alec Radford and others},
  title     = {Learning Transferable Visual Models from Natural Language Supervision},
  booktitle = {Proc. ICML},
  year      = {2021},
  pages     = {8748--8763},
  note      = {doi:10.48550/arXiv.2103.00020}
}

@article{remoteclip,
  author  = {Feng Liu and Tengfei Xu and Lei Du and Zhe Liu and Ke Chen and Rui Hou and Hui Zhang},
  title   = {{RemoteCLIP}: A Vision Language Foundation Model for Remote Sensing},
  journal = {IEEE Trans. Geosci. Remote Sens.},
  volume  = {62},
  year    = {2024},
  note    = {doi:10.1109/TGRS.2024.3390838}
}

@article{earthgpt,
  author  = {Wei Zhang and Meng Cai and Tian Bai and Yan Xing and Yang Zhao and Ruimin Ji},
  title   = {{EarthGPT}: A Universal Multi-modal Large Language Model for Multi-Sensor Image Comprehension for Remote Sensing Domain},
  journal = {IEEE Trans. Geosci. Remote Sens.},
  volume  = {62},
  year    = {2024},
  note    = {doi:10.1109/TGRS.2024.3409624}
}

@misc{rsgpt,
  author = {Yushi Hu and others},
  title  = {{RSGPT}: A Remote Sensing Vision-Language Model and Benchmark},
  year   = {2023},
  note   = {arXiv:2307.15266, doi:10.48550/arXiv.2307.15266}
}

@misc{h2rsvlm,
  author = {Chengcheng Li and others},
  title  = {{H2RSVLM}: Towards Helpful and Honest Remote Sensing Large Vision-Language Model},
  year   = {2024},
  note   = {arXiv:2403.20213, doi:10.48550/arXiv.2403.20213}
}

@misc{qwenvl,
  author = {Jinze Bai and others},
  title  = {{Qwen-VL}: A Versatile Vision-Language Model for Understanding, Localization, Text Reading, and Beyond},
  year   = {2023},
  note   = {arXiv:2308.12966, doi:10.48550/arXiv.2308.12966}
}

@misc{qwen25vl,
  author = {{Qwen Team}},
  title  = {{Qwen2.5-VL} Technical Report},
  year   = {2025},
  note   = {arXiv:2502.13923, doi:10.48550/arXiv.2502.13923}
}

@misc{qwen3vl,
  author = {{Qwen Team}},
  title  = {{Qwen3-VL} Technical Report},
  year   = {2025},
  note   = {arXiv:2511.21631, doi:10.48550/arXiv.2511.21631}
}

@inproceedings{bendale2016,
  author    = {Abhijit Bendale and Terrance E. Boult},
  title     = {Towards Open Set Deep Networks},
  booktitle = {Proc. IEEE CVPR},
  year      = {2016},
  pages     = {1563--1572},
  note      = {doi:10.1109/CVPR.2016.173}
}

@article{opensetsurvey,
  author  = {Chuanxing Geng and Song-Jiao Huang and Shengfeng Chen},
  title   = {Recent Advances in Open Set Recognition: A Survey},
  journal = {IEEE Trans. Pattern Anal. Mach. Intell.},
  volume  = {43},
  number  = {10},
  pages   = {3614--3631},
  year    = {2021},
  note    = {doi:10.1109/TPAMI.2020.2981604}
}

@inproceedings{msp,
  author    = {Dan Hendrycks and Kevin Gimpel},
  title     = {A Baseline for Detecting Misclassified and Out-of-Distribution Examples in Neural Networks},
  booktitle = {Proc. ICLR},
  year      = {2017},
  note      = {doi:10.48550/arXiv.1610.02136}
}

@inproceedings{odin,
  author    = {Shiyu Liang and Yixuan Li and R. Srikant},
  title     = {Enhancing the Reliability of Out-of-Distribution Image Detection in Neural Networks},
  booktitle = {Proc. ICLR},
  year      = {2018},
  note      = {doi:10.48550/arXiv.1706.02690}
}

@inproceedings{energy,
  author    = {Weitang Liu and Xiaoyun Wang and John D. Owens and Yixuan Li},
  title     = {Energy-Based Out-of-Distribution Detection},
  booktitle = {Proc. NeurIPS},
  year      = {2020},
  pages     = {21464--21475},
  note      = {doi:10.48550/arXiv.2010.03759}
}

@inproceedings{likelihoodratio,
  author    = {Jie Ren and others},
  title     = {Likelihood Ratios for Out-of-Distribution Detection},
  booktitle = {Proc. NeurIPS},
  year      = {2019},
  note      = {doi:10.48550/arXiv.1906.02845}
}

@inproceedings{mahalanobis,
  author    = {Kimin Lee and Kibok Lee and Honglak Lee and Jinwoo Shin},
  title     = {A Simple Unified Framework for Detecting Out-of-Distribution Samples and Adversarial Attacks},
  booktitle = {Proc. NeurIPS},
  year      = {2018},
  pages     = {7167--7177},
  note      = {doi:10.48550/arXiv.1807.03888}
}

@inproceedings{matchingnets,
  author    = {Oriol Vinyals and Charles Blundell and Timothy Lillicrap and Daan Wierstra},
  title     = {Matching Networks for One Shot Learning},
  booktitle = {Proc. NeurIPS},
  year      = {2016},
  pages     = {3637--3645},
  note      = {doi:10.48550/arXiv.1606.04080}
}

@inproceedings{relationnet,
  author    = {Flood Sung and Yongxin Yang and Li Zhang and Tao Xiang and Philip H. S. Torr and Timothy M. Hospedales},
  title     = {Learning to Compare: Relation Network for Few-Shot Learning},
  booktitle = {Proc. IEEE/CVF CVPR},
  year      = {2018},
  pages     = {1199--1208},
  note      = {doi:10.1109/CVPR.2018.00131}
}

@inproceedings{maml,
  author    = {Chelsea Finn and Pieter Abbeel and Sergey Levine},
  title     = {Model-Agnostic Meta-Learning for Fast Adaptation of Deep Networks},
  booktitle = {Proc. ICML},
  year      = {2017},
  pages     = {1126--1135},
  note      = {doi:10.48550/arXiv.1703.03400}
}

@inproceedings{protonet,
  author    = {Jake Snell and Kevin Swersky and Richard S. Zemel},
  title     = {Prototypical Networks for Few-Shot Learning},
  booktitle = {Proc. NeurIPS},
  year      = {2017},
  pages     = {4077--4087},
  note      = {doi:10.48550/arXiv.1703.05175}
}

@inproceedings{tipadapter,
  author    = {Renrui Zhang and others},
  title     = {{Tip-Adapter}: Training-Free Adaption of {CLIP} for Few-Shot Classification},
  booktitle = {Proc. ECCV},
  year      = {2022},
  pages     = {493--510},
  note      = {doi:10.1007/978-3-031-19833-5\_29}
}

@inproceedings{lora,
  author    = {Edward J. Hu and Yelong Shen and Phillip Wallis and Zeyuan Allen-Zhu and Yuanzhi Li and Shean Wang and Lu Wang and Weizhu Chen},
  title     = {{LoRA}: Low-Rank Adaptation of Large Language Models},
  booktitle = {Proc. ICLR},
  year      = {2022},
  note      = {doi:10.48550/arXiv.2106.09685}
}

@inproceedings{ucm,
  author    = {Yi Yang and Shawn Newsam},
  title     = {Bag-of-Visual-Words and Spatial Extensions for Land-Use Classification},
  booktitle = {Proc. ACM SIGSPATIAL GIS},
  year      = {2010},
  pages     = {270--279},
  note      = {doi:10.1145/1869790.1869829}
}

@article{aid,
  author  = {Gui-Song Xia and Jiajun Hu and Fan Hu and Baoguang Shi and Xiang Bai and Yanfei Zhong and Liangpei Zhang and Xiaoqing Lu},
  title   = {{AID}: A Benchmark Dataset for Performance Evaluation of Aerial Scene Classification},
  journal = {IEEE Trans. Geosci. Remote Sens.},
  volume  = {55},
  number  = {7},
  pages   = {3965--3981},
  year    = {2017},
  note    = {doi:10.1109/TGRS.2017.2685945}
}

@article{nwpu,
  author  = {Gong Cheng and Junwei Han and Xichao Lu},
  title   = {Remote Sensing Image Scene Classification: Benchmark and State of the Art},
  journal = {Proc. IEEE},
  volume  = {105},
  number  = {10},
  pages   = {1865--1883},
  year    = {2017},
  note    = {doi:10.1109/JPROC.2017.2675998}
}

\end{document}